\documentclass[conference]{IEEEtran}
\IEEEoverridecommandlockouts
\usepackage{cite}
\usepackage{amsmath,amssymb,amsfonts}
\usepackage{algorithmic}
\usepackage{graphicx}
\usepackage{textcomp}
\usepackage{xcolor}
\usepackage[caption=false,font=footnotesize]{subfig}
\usepackage{url}
\usepackage{booktabs}
\usepackage{makecell}
\def\BibTeX{{\rm B\kern-.05em{\sc i\kern-.025em b}\kern-.08em
    T\kern-.1667em\lower.7ex\hbox{E}\kern-.125emX}}
\begin{document}

\title{Fine-Grained Food Image Understanding via Target-Aware Data Alignment
}

\author{\IEEEauthorblockN{Jui-Feng Chi\textsuperscript{*}, Wei-Lun Chu\textsuperscript{*}, Bruce Coburn, Jinge Ma, Fengqing Zhu}

\IEEEauthorblockA{
\textit{\textsuperscript{*}Equal contribution.}\\
\textit{Elmore Family School of Electrical and Computer Engineering}\\
Purdue University, West Lafayette, IN, USA}
}

\maketitle

\begin{abstract}
Fine-grained food visual--semantic understanding requires models to capture subtle distinctions across ingredients, cooking methods, doneness, color, texture, and plate composition. Although CLIP-style vision-language models provide a natural framework for this task, their effectiveness is limited when training relies on heterogeneous web-collected image--text pairs. Such data often exhibit a web-to-target domain gap and cross-modal misalignment, where images differ from the target distribution and captions are noisy, multilingual, or weakly grounded in visual content. We propose a data-centric multimodal alignment method for fine-grained food description and recognition. Our method first performs target-aware data selection to identify visually relevant training subsets, then applies VLM-based caption refinement to generate visually grounded, target-style descriptions. Using these curated image--caption pairs, we train complementary CLIP-style retrieval experts and further combine their decisions through a hierarchical VLM-assisted multi-expert decision-level fusion strategy that invokes the VLM only when experts disagree. Experiments show that our data refinement strategy significantly improves retrieval performance over naive web supervision, with VLM-based caption refinement alone yielding an average performance gain of approximately 19\%. Our full method also achieves more than twice the retrieval score of pure VLM-based retrieval while remaining substantially more efficient.
\end{abstract}


\section{Introduction}

Understanding food semantics from images is a particularly challenging computer vision problem~\cite{min_2023}. Unlike generic object recognition, food image understanding often requires models to capture fine-grained details across food categories, ingredient composition, cooking methods, doneness, color, texture, and plate composition. For example, two images may both contain chicken and vegetables, but one may show grilled chicken with charred textures served with roasted carrots, while the other may depict tender pan-seared chicken coated in sauce and served with saut\'{e}ed broccoli. Their differences are not limited to food categories, but also lie in fine-grained attributes such as cooking method, ingredient state, and visual texture. These distinctions are particularly important for dietary monitoring, nutritional analysis, and food-safety applications, where systems must go beyond the capabilities of conventional food classification models~\cite{bossard_2014}.

\begin{figure*}[!t]
\centering
\subfloat[Image-space source-to-target domain gap.]{
    \includegraphics[width=0.275\textwidth]{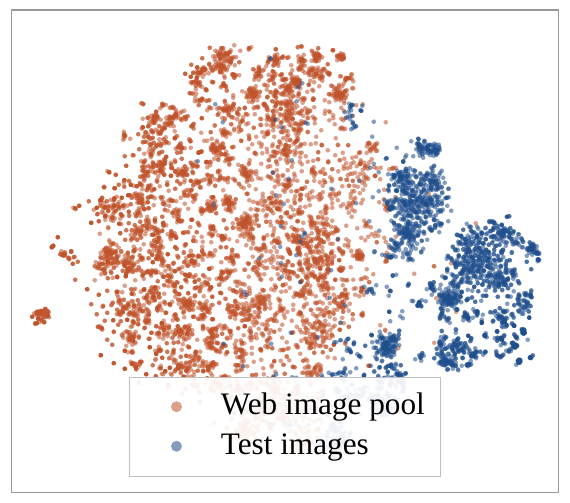}
    \label{fig:intro_image_before}
}
\hfill
\subfloat[Text-space source-to-target domain gap with raw web captions.]{
    \includegraphics[width=0.275\textwidth]{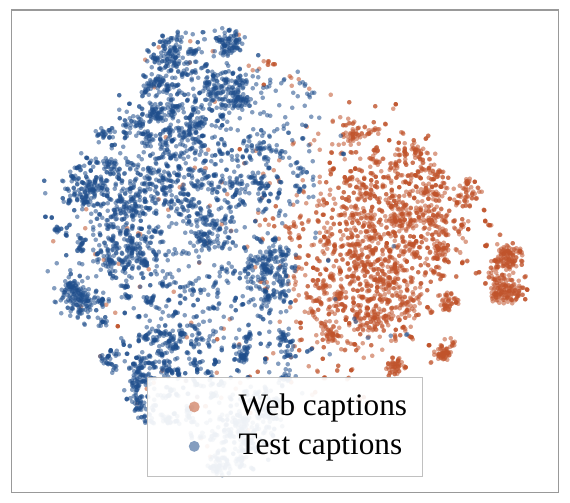}
    \label{fig:intro_text_before}
}
\hfill
\subfloat[Test target pair vs. web training pair.]{
    \includegraphics[width=0.38\textwidth]{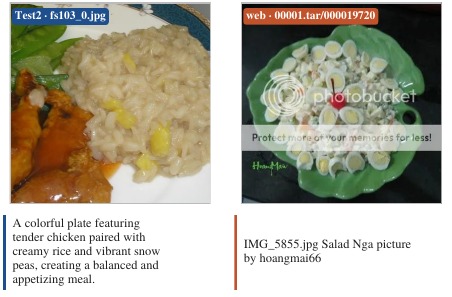}
    \label{fig:intro_pair_web}
}
\caption{
t-SNE visualization of the source-to-target domain gap in the available training data, illustrated using Dishcovery Mission II challenge~\cite{radeva_2026}. The example pair in (c) further shows that web captions can be weakly aligned with the visible food content.
}
\label{fig:intro_before_selection}
\end{figure*}

Contrastive vision-language models, such as CLIP~\cite{radford_2021} and its successors~\cite{jia_2021, ilharco_2021, zhai_2023}, provide a natural solution for fine-grained food image--text retrieval by mapping food images and fixed or expressive textual descriptions into a shared embedding space. However, such fine-grained cross-modal alignment is limited not only by model capacity, but also by the quality of large-scale training data. For food image analysis, large-scale training data often consists of web-collected image--text pairs, where both image and text modalities are highly heterogeneous: images may come from recipe websites, menus, or social media, with diverse shooting conditions and presentation styles; texts may be multilingual dish names, user tags, menu descriptions, or website comments. While this diversity enables data scaling, it also introduces substantial noise, cross-modal misalignment, and a source-to-target domain gap between the available web training data and the target data distribution. Fig.~\ref{fig:intro_before_selection} illustrates this in both image and text spaces. Without selecting, cleaning, and improving the semantic alignment between images and texts, directly training on raw image--text pairs can make it difficult for models to acquire fine-grained visual--semantic understanding and may even degrade retrieval performance due to noisy supervision.

We argue that when training data exhibits substantial heterogeneity and alignment bias, improving data quality is more critical than solely modifying model architectures or training objectives. We distill this intuition into a three-stage data-centric multimodal alignment strategy for fine-grained food description and recognition:

\begin{enumerate}

\item \textbf{Target-aware data selection}: We rank web images by their proximity to the target distribution and select compact, relevant subsets. We show that a strategically selected 5k subset can outperform a naively selected 25k subset under raw web supervision, challenging the assumption that more data is always better even when the data are noisy.

\item \textbf{VLM-based caption refinement}: We replace noisy web captions with visually grounded descriptions generated by a vision-language model, providing richer training supervision without collecting or generating new images. This single step yields a roughly 19\% performance gain and is even more effective than the choice of backbone or fine-tuning strategy.

\item \textbf{Hierarchical multi-expert decision-level fusion}: We fuse decisions from complementary experts through VLM-assisted disagreement resolution, invoking the VLM only when their predictions disagree.

\end{enumerate}

\section{Related Work}
\subsection{CLIP-Based Image-Text Retrieval}

CLIP~\cite{radford_2021} learns a shared image--text embedding space through large-scale contrastive pretraining. ALIGN~\cite{jia_2021} scales contrastive vision-language learning to noisy web data, and OpenCLIP~\cite{ilharco_2021} enables open-source CLIP training at scale. However, mismatches between web-collected training pairs and target data can still limit retrieval performance. We address this issue by selecting target-relevant web images and refining their captions to better match the target style.

\subsection{Food Recognition and Food Retrieval}

Food recognition is a fine-grained problem because food images often contain multiple ingredients, sauces, side dishes, and visually similar cooking styles. Food-101~\cite{bossard_2014} supports large-scale food classification, while Recipe1M+~\cite{marin_2021} connects food images with structured recipe information for cross-modal retrieval.

Recent food retrieval tasks require finer alignment between images and descriptive text. Unlike conventional classification and recipe retrieval, Dishcovery Mission II~\cite{radeva_2026} evaluates multi-ingredient recognition and single-caption retrieval. The latter matches an image to a detailed caption describing ingredients, side items, textures, colors, sauces, and plate composition. Our work follows this fine-grained food image-to-text retrieval setting.

\subsection{Decision-Level Expert Fusion and Reranking}

Reranking and rank fusion are common strategies for improving retrieval results by refining or combining initial predictions. Neural rerankers use stronger models to rescore retrieved candidates~\cite{nogueira_2019}, while rank fusion methods combine outputs from multiple retrieval systems to exploit complementary strengths~\cite{cormack_2009}. Recent work has also explored large language models as zero-shot reranking agents for information retrieval, where they reorder candidate results according to their relevance to a query~\cite{sun_2023}. Inspired by this line of work, we use a VLM to resolve prediction disagreements among complementary CLIP-style experts.

\begin{figure*}[t]
    \centering
    \includegraphics[
        width=0.95\textwidth,
        trim=0 0.35in 0 0.5in,
        clip
    ]{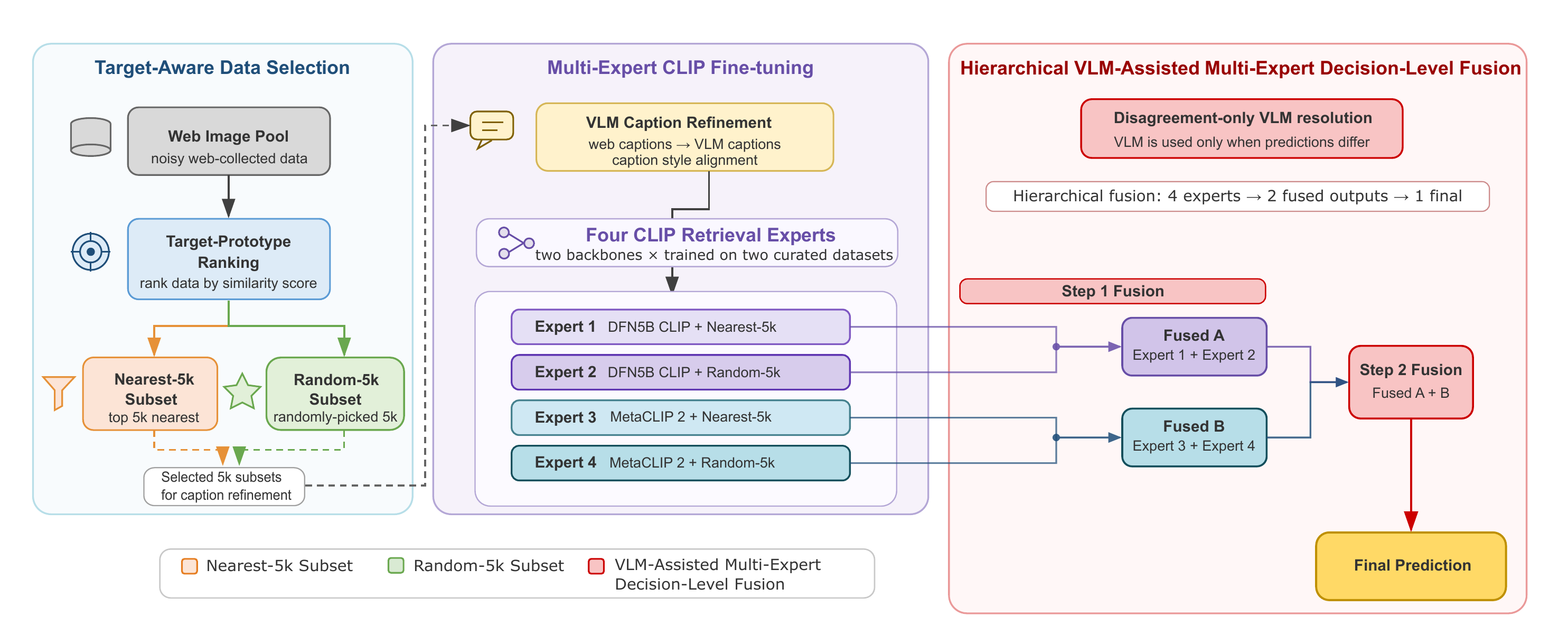}
    \caption{Overview of the proposed three-stage data-centric food image–text retrieval framework.}
    \label{fig:flowchart}
\end{figure*}

\section{Methods}

\subsection{Overview}

We propose a three-stage data-centric multimodal alignment method for fine-grained food visual-semantic understanding. The method is designed to reduce both the source-to-target visual distribution gap and the image--text semantic misalignment in web-collected training data. In this work, we instantiate it in a food image-to-text retrieval setting. As shown in Fig.~\ref{fig:flowchart}, the method first selects compact target-relevant subsets from noisy web-collected image--caption pairs, then regenerates their captions to better match the target evaluation style. Finally, multiple CLIP-style retrieval experts trained from different curated subsets and backbones are combined through hierarchical VLM-assisted decision-level fusion, where the VLM is invoked only when expert predictions disagree.

\subsection{Target-Aware Data Selection}
Our goal is to select web images that are visually closer to the test-time retrieval distribution. We therefore summarize the target visual distribution with an image prototype and rank web images by their similarity to this prototype.

Let $g_I(\cdot)$ denote a fixed pretrained CLIP image encoder used only for data selection. Here, the target set refers to the unlabeled test images. 
Given these images $\{x_j^{\mathrm{tar}}\}_{j=1}^{N}$, we compute an average normalized target prototype:
\begin{equation}
    p_{\mathrm{tar}} =
    \frac{1}{N}
    \sum_{j=1}^{N}
    \frac{g_I(x_j^{\mathrm{tar}})}
    {\|g_I(x_j^{\mathrm{tar}})\|_2}.
\end{equation}
Thus, $p_{\mathrm{tar}}$ represents the centroid of the normalized target image embeddings in the CLIP feature space. For each web image $x_i^{\mathrm{web}}$, we compute its similarity to the target prototype:
\begin{equation}
    r_i =
    \left(
    \frac{g_I(x_i^{\mathrm{web}})}
    {\|g_I(x_i^{\mathrm{web}})\|_2}
    \right)^\top
    \frac{p_{\mathrm{tar}}}
    {\|p_{\mathrm{tar}}\|_2}.
\end{equation}
Since both vectors in Eq.~(2) are $\ell_2$-normalized, $r_i$ is their cosine similarity and serves as a scalar target-alignment score. A higher $r_i$ indicates that the corresponding web image is more closely aligned with the target prototype. We rank all web images in descending order of $r_i$ and retain the top 5k as the target-aware subset, denoted as \textbf{Nearest-5k}. This subset is designed to reduce the source-to-target visual distribution gap by selecting web images that are closest to the target image distribution.

In addition, we construct a second subset, denoted as \textbf{Random-5k}, by randomly sampling 5k images from the available web image pool. Unlike Nearest-5k, Random-5k does not explicitly optimize for target similarity. Instead, it preserves broader diversity from the original web data distribution, which can lead to different expert prediction patterns and provide complementary signals during later multi-expert decision-level fusion.

\subsection{VLM-Based Caption Refinement}

\begin{table*}[!t]
\centering
\caption{The Effects of Data Selection and Caption Refinement Across Fine-Tuning Methods}
\label{tab:adaptation_methods}
\setlength{\tabcolsep}{3.8pt}
\renewcommand{\arraystretch}{1.12}
\footnotesize
\begin{tabular}{@{}lccccccc@{}}
\toprule
\textbf{Data and Caption Setting}
& \textbf{Full FT}
& \shortstack{\textbf{Last Few}\\\textbf{Layers FT}}
& \shortstack{\textbf{CLIP}\\\textbf{Adapter}~\cite{gao_2024}}
& \textbf{LoRA}~\cite{hu_2022}
& \textbf{DoRA}~\cite{liu_2024}
& \textbf{NegCLIP}~\cite{yuksekgonul_2023}
& \textbf{WiSE-FT}~\cite{wortsman_2022} \\
\midrule

\multicolumn{8}{@{}l}{\textit{Backbone: DFN5B CLIP~\cite{fang_2024} \quad (Pretrained: 0.509)}} \\
Random-5k + Raw Cap.
& 0.464 & 0.500 & 0.424 & 0.380 & 0.375 & 0.440 & 0.504 \\
Random-25k + Raw Cap.
& 0.380 & 0.443 & 0.332 & 0.324 & 0.320 & 0.380 & 0.478 \\
Nearest-5k + Raw Cap.
& 0.477 & 0.506 & 0.431 & 0.412 & 0.413 & 0.484 & 0.514 \\
Random-5k + VLM Cap.
& \textbf{0.592} & 0.599 & 0.455 & 0.565 & 0.565 & \textbf{0.497} & \textbf{0.580} \\
Nearest-5k + VLM Cap.
& 0.589 & \textbf{0.607} & \textbf{0.476} & \textbf{0.574} & \textbf{0.575} & 0.457 & 0.570 \\
\midrule

\multicolumn{8}{@{}l}{\textit{Backbone: MetaCLIP 2~\cite{chuang_2025} \quad (Pretrained: 0.450)}} \\
Random-5k + Raw Cap.
& 0.466 & 0.474 & 0.441 & 0.475 & 0.471 & 0.475 & 0.459 \\
Random-25k + Raw Cap.
& 0.490 & 0.443 & 0.415 & 0.474 & 0.477 & 0.449 & 0.478 \\
Nearest-5k + Raw Cap.
& 0.498 & 0.496 & 0.453 & 0.489 & 0.489 & 0.480 & 0.470 \\
Random-5k + VLM Cap.
& 0.584 & 0.574 & 0.485 & \textbf{0.588} & \textbf{0.587} & \textbf{0.509} & \textbf{0.550} \\
Nearest-5k + VLM Cap.
& \textbf{0.587} & \textbf{0.575} & \textbf{0.486} & 0.583 & 0.585 & 0.493 & 0.537 \\
\bottomrule
\end{tabular}
\end{table*}

Because raw web captions are often short, noisy, or weakly aligned with the visible food content, we regenerate captions for the selected images using a VLM. 
Specifically, we use Gemma 4~\cite{google_deepmind_2026} to generate one English caption for each selected web image. 
To make the generated captions closer to the target evaluation style, we randomly sample unpaired target captions as in-context examples and prompt the VLM to describe only observable food details, such as visible ingredients, side dishes, sauces, textures, colors, and plate composition. 
The prompt constrains the output to a concise food caption without bullet points, explanations, or irrelevant non-food details. 
We then apply simple text cleaning to remove common prefixes, quotation marks, line breaks, and formatting artifacts. 
This produces curated image--caption pairs $(x_i^{\mathrm{web}}, \hat{t}_i^{\mathrm{vlm}})$ for retrieval expert training.

\subsection{Retrieval Expert Construction}

We construct retrieval experts using two pretrained CLIP-style backbones: 
DFN5B-CLIP-ViT-H-14~\cite{fang_2024}, hereafter referred to as DFN5B, and MetaCLIP~2~\cite{chuang_2025}. Each backbone is trained on both curated subsets, producing four experts in total. This results in four retrieval experts:
\begin{itemize}
    \item $E_1$: DFN5B trained on Nearest-5k.
    \item $E_2$: DFN5B trained on Random-5k.
    \item $E_3$: MetaCLIP~2 trained on Nearest-5k.
    \item $E_4$: MetaCLIP~2 trained on Random-5k.
\end{itemize}

Each expert is trained with the standard CLIP contrastive objective, implemented as a symmetric InfoNCE loss~\cite{radford_2021,van_den_oord_2018}. 
This stage simply produces retrieval experts with different backbones and curated datasets, rather than introducing a new training objective.

At inference time, each expert independently predicts retrieval results. 
For single-caption retrieval, the expert selects the candidate caption with the highest image--text similarity. 
For multi-ingredient recognition, it ranks ingredient labels by image--text similarity and selects labels using thresholding.

\subsection{Hierarchical VLM-Assisted Multi-Expert Decision-Level Fusion}

The four experts may produce different predictions because they are trained on different curated datasets and pretrained backbones. 
Instead of merging all outputs by simple voting or score averaging, we use a VLM to resolve prediction disagreements.

Fusion follows a simple disagreement-based rule: if two expert predictions agree, we keep the shared prediction; otherwise, the VLM reviews the image and the disagreement to make a visual judgment. 
For single-caption retrieval, the disagreement is represented as two candidate captions, and the VLM selects the caption that better matches the image. 
For multi-label ingredient retrieval, common ingredients are kept unchanged, and disputed ingredients are resolved one by one through visual yes/no questions, which is more efficient than asking the VLM to predict the full ingredient set. 
Thus, the VLM is used selectively for decision-level fusion between experts rather than as a full retrieval model.

We apply this rule hierarchically to the outputs of the four experts defined above. 
First, the predictions from $E_1$ and $E_2$ are fused into a DFN5B branch output $M_{\mathrm{DFN}}$, while the predictions from $E_3$ and $E_4$ are fused into a MetaCLIP~2 branch output $M_{\mathrm{Meta}}$. 
Second, $M_{\mathrm{DFN}}$ and $M_{\mathrm{Meta}}$ are fused with the same VLM-assisted rule to obtain the final output.

\section{Results}
\subsection{Experimental Setup}

We evaluate the proposed method on the official Dishcovery Mission II challenge test set~\cite{radeva_2026}, which combines a multi-ingredient recognition task and a descriptive caption retrieval task. Following the challenge protocol, we report the official score, defined as the harmonic mean of the multi-ingredient recognition F1-score and the descriptive caption retrieval accuracy.

For each potential retrieval expert, we compare seven CLIP training
strategies to select a strong checkpoint: full fine-tuning,
last-few-layers fine-tuning, CLIP-Adapter~\cite{gao_2024}, LoRA~\cite{hu_2022}, DoRA~\cite{liu_2024}, NegCLIP~\cite{yuksekgonul_2023}, and
WiSE-FT~\cite{wortsman_2022}.
Unless otherwise specified, all settings use the same training budget:
6 epochs, batch size 128, warmup ratio 0.1, and AdamW with cosine decay. For the last-few-layers fine-tuning setting, both DFN5B and MetaCLIP~2 unfreeze the last 8 layers of the vision encoder and the last 4 layers of the text encoder. All pretrained model weights are initialized from publicly available checkpoints hosted on the Hugging Face Hub~\cite{hugging_face_2026}.
Backbone-specific learning rates and unfrozen-layer configurations are kept
consistent across data subsets to ensure fair comparison.

\subsection{Effect of Data and Caption Refinement}

Table~\ref{tab:adaptation_methods} summarizes the effects of data selection and caption refinement across different fine-tuning methods and two CLIP-style backbones. The strongest overall result is achieved by DFN5B with last-few-layers fine-tuning on Nearest-5k + VLM Cap., reaching 0.607. More importantly, the comparison shows that caption refinement is critical: for DFN5B, fine-tuning with raw web captions often degrades performance compared with the pretrained model, whose zero-shot score is 0.509. Even the best raw-caption setting only slightly exceeds the pretrained DFN5B score, and increasing the raw web data to 25k samples still fails to match the VLM-captioned 5k settings. Although MetaCLIP~2 does not show the same degradation relative to its pretrained score, it follows a similar overall trend, where raw-caption fine-tuning remains limited and VLM-captioned settings are substantially stronger. This suggests that noisy and weakly aligned web captions, rather than data scale alone, are a major bottleneck for food image--text retrieval. Overall, VLM caption refinement improves model performance by approximately 19\% on average.

The effect of data selection is also clear before caption refinement. Under raw-caption settings, Nearest-5k consistently outperforms Random-5k across all fine-tuning methods and both backbones, indicating that target-aware image selection helps reduce the visual domain gap. However, after VLM caption refinement, this trend becomes less uniform: the relative benefit of Nearest-5k depends on the backbone and fine-tuning method. This suggests that once captions are rewritten into a more visually grounded form, the remaining bottleneck shifts from data similarity alone toward the quality of VLM caption refinement and its interaction with each fine-tuning strategy. 

\begin{table}[!t]
\centering
\caption{Performance and Efficiency Comparison with pure VLM Retrieval.
Inference time is measured on $2\times$H100 GPUs for Pure Gemma~4 and on a GB10 machine for our full method.}
\label{tab:pipeline_vs_vlm}
\setlength{\tabcolsep}{8pt}
\renewcommand{\arraystretch}{1.3}
\begin{tabular}{@{}l c c c c@{}}
\toprule
\textbf{Method}
 & \textbf{Score}
 & \begin{tabular}{@{}c@{}}VLM\\calls\end{tabular}
 & \begin{tabular}{@{}c@{}}VLM tokens\\(M)\end{tabular}
 & \begin{tabular}{@{}c@{}}Inf. time\\(min)\end{tabular} \\
\midrule
Pure Gemma\,4~\cite{google_deepmind_2026}
& 0.226 & 24,107 & 431.2 & 1760.4 \\
\textbf{Ours: full method}
& \textbf{0.653} & \textbf{15,235} & \textbf{6.1} & \textbf{247.6} \\
\bottomrule
\end{tabular}
\end{table}
\begin{table}[!t]
\centering
\caption{Ablation of Training Data Selection with VLM-Generated Captions}
\label{tab:data_selection_ablation}
\setlength{\tabcolsep}{2.0pt}
\renewcommand{\arraystretch}{1.08}
\footnotesize
\begin{tabular}{@{}l l c c@{}}
\toprule
\textbf{Subset}
& \textbf{Selection Criterion}
& \makecell[c]{DFN5B\\CLIP}
& \makecell[c]{MetaCLIP~2} \\
\midrule
Random-5k
& Random sampling
& 0.599
& 0.584 \\
Nearest-25k
& Top visual similarity
& 0.581
& 0.584 \\
Nearest-5k
& Top visual similarity
& \textbf{0.607}
& \textbf{0.587} \\
\bottomrule
\end{tabular}
\end{table}
\subsection{Complete Method Performance}

Table~\ref{tab:pipeline_vs_vlm} compares our full method with directly using Gemma~4 as a pure zero-shot retrieval model. For the pure VLM baseline, ideally all candidate captions would be provided in a single prompt for retrieval. However, doing so exceeds practical GPU memory limits due to the extremely long context length, so we instead split the candidates into chunks of at most 1,000 captions and let Gemma~4 select the best match within each chunk before performing a merge step for the final prediction. Even with this memory-aware strategy, direct VLM retrieval remains expensive and ineffective, achieving only 0.226 while requiring 24,107 VLM calls and 431.2M tokens.

In contrast, our full method uses Gemma~4 selectively. Following the expert construction described earlier, we instantiate four VLM-caption-trained retrieval experts, $E_1$--$E_4$, using the selected adaptation setting for each backbone in Table~\ref{tab:adaptation_methods}: last-few-layers fine-tuning for DFN5B and full fine-tuning for MetaCLIP~2. Since full fine-tuning and LoRA show comparable performance on MetaCLIP~2, we select full fine-tuning for its simpler and more reproducible setup. With these adaptation settings, the decision-level fusion of the four experts achieves 0.653, improving over direct Gemma~4 retrieval by 0.427 absolute points, while reducing token usage from 431.2M to 6.1M. These results show that targeted VLM usage is both more accurate and more token-efficient than using the VLM as a full retrieval model.

For inference time, we report the measured runtime and hardware for reference. Since runtime is hardware-dependent, we primarily use VLM calls and token counts as hardware-independent efficiency indicators.

\subsection{Ablation Study}
\begin{figure}[!t]
\centering
\includegraphics[width=\columnwidth]{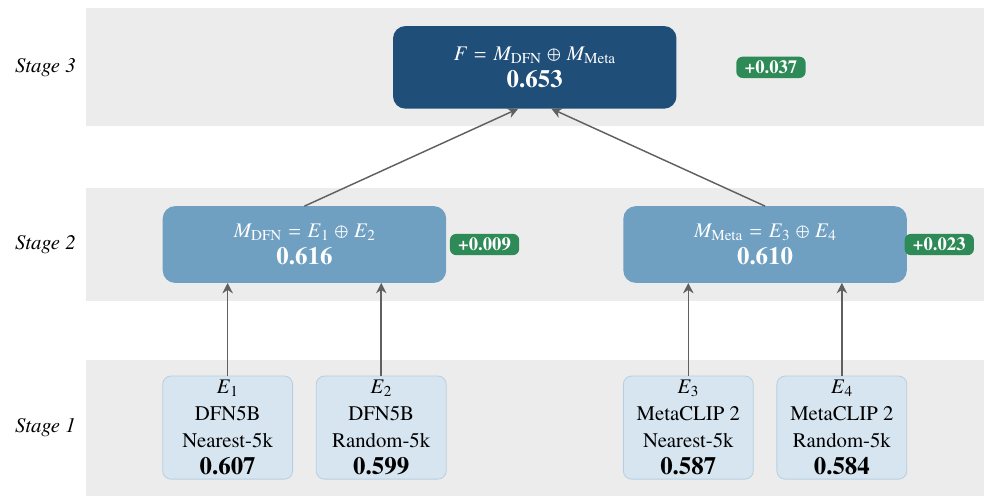}
\caption{
Hierarchical VLM-assisted multi-expert decision-level fusion.
}
\label{fig:hierarchical_merge}
\end{figure}

Fig.~\ref{fig:hierarchical_merge} illustrates the proposed hierarchical VLM-assisted decision-level fusion process. The best single expert, $E_1$, achieves a score of 0.607. By combining the predictions of the two DFN5B experts, the score improves to 0.616, while applying the same fusion strategy to the MetaCLIP~2 experts yields 0.610. The final cross-backbone fusion between the DFN5B and MetaCLIP~2 branches further increases the score to 0.653. This progression shows that VLM-assisted disagreement resolution can effectively integrate complementary expert decisions and improve beyond the strongest individual expert.

Table~\ref{tab:data_selection_ablation} isolates the effect of training data selection when all settings use VLM-generated captions. We use the same tuning settings as the final retrieval experts: last-few-layers fine-tuning for DFN5B and full fine-tuning for MetaCLIP~2. Nearest-5k achieves the best result for both backbones, reaching 0.607 and 0.587, respectively. Increasing the selected data to 25k does not further improve performance, suggesting that once captions are refined, a compact and target-aligned 5k subset already provides sufficient supervision. 

\subsection{Qualitative Analysis}
\begin{figure*}[!t]
\centering
\subfloat[Image-space distribution after Nearest-5k selection.]{
    \includegraphics[width=0.275\textwidth]{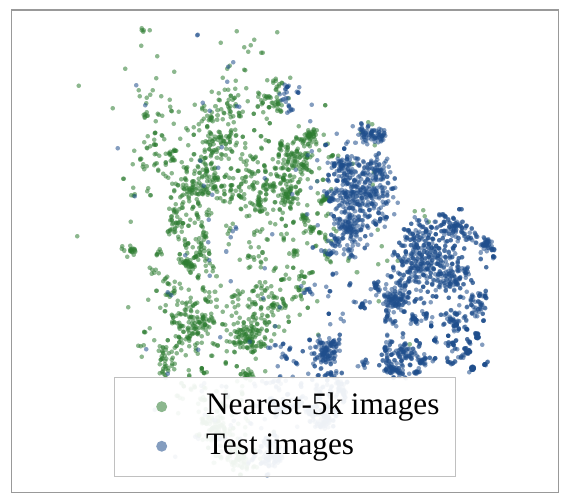}
    \label{fig:result_image_after}
}
\hfill
\subfloat[Text-space distribution after VLM caption refinement.]{
    \includegraphics[width=0.275\textwidth]{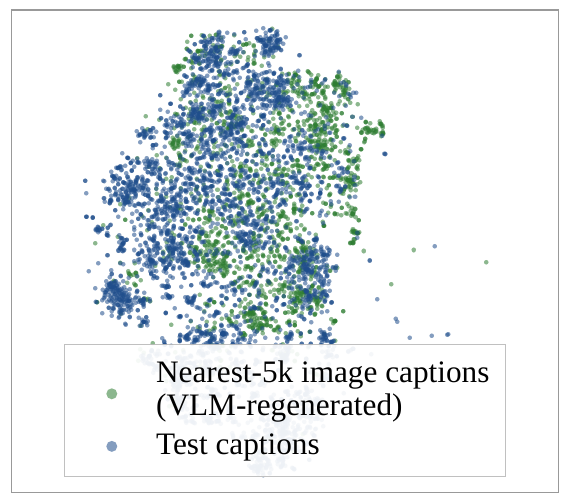}
    \label{fig:result_text_after}
}
\hfill
\subfloat[Example alignment between a target pair and a VLM-captioned Nearest-5k pair.]{
    \includegraphics[width=0.38\textwidth]{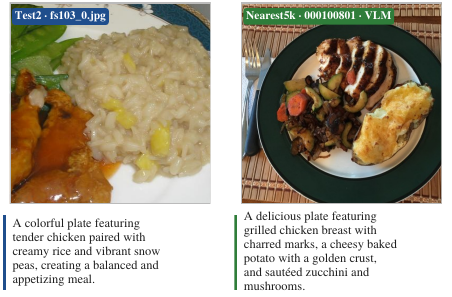}
    \label{fig:result_pair_nearest5k}
}
\caption{
Effect of data selection and caption refinement. Nearest-5k selection narrows the visual domain gap, while VLM-generated captions reduce the text-space distribution gap. Panel (c) shows an aligned training pair after both stages.
}
\label{fig:result_after_selection}
\end{figure*}

Fig.~\ref{fig:result_after_selection} provides a qualitative view of the proposed data refinement method. 
After Nearest-5k selection, the selected web images become closer to the Test 2 image distribution, reducing the source-to-target visual distribution gap observed before selection. 
After VLM-based caption refinement, the regenerated captions largely overlap with the target caption distribution in the t-SNE visualization, suggesting that caption refinement effectively reduces the text-space distribution gap.

The example pair in Fig.~\ref{fig:result_after_selection} further illustrates this effect. 
Compared with the raw web example in Fig.~\ref{fig:intro_before_selection}, the selected Nearest-5k sample contains a clearer food image and a caption that describes visible ingredients, textures, and plate composition. 
This shows that our method improves both source-to-target visual relevance and the image--text semantic misalignment.

\section{Conclusion}

We presented a three-stage data-centric multimodal alignment method for fine-grained food understanding under weak web supervision. Our method improves heterogeneous web-collected image--text pairs through target-aware data selection, VLM-based caption refinement, and complementary multi-expert decision-level fusion, producing training pairs that are more visually relevant and semantically grounded. Experiments on both multi-ingredient recognition and single-caption retrieval show that caption refinement yields an average gain of approximately 19\%. By combining complementary retrieval experts with hierarchical VLM-assisted multi-expert decision-level fusion, our full method achieves more than twice the image-to-text retrieval score of pure VLM-based retrieval while requiring substantially lower VLM resource usage. These findings highlight the importance of principled data curation and selective VLM usage when training data is noisy, heterogeneous, or cross-modally misaligned.

\bibliographystyle{IEEEtran}
\bibliography{references}

@article{radford_2021,
  title     = {Learning Transferable Visual Models From Natural Language Supervision},
  author    = {Radford, Alec and Kim, Jong Wook and Hallacy, Chris and Ramesh, Aditya and Goh, Gabriel and Agarwal, Sandhini and Sastry, Girish and Askell, Amanda and Mishkin, Pamela and Clark, Jack and Krueger, Gretchen and Sutskever, Ilya},
  journal   = {Proceedings of the 38th International Conference on Machine Learning},
  volume    = {139},
  series    = {Proceedings of Machine Learning Research},
  pages     = {8748--8763},
  year      = {2021},
  month     = {July},
  note      = {{Virtual Event}},
  publisher = {PMLR},
  url       = {http://proceedings.mlr.press/v139/radford21a.html}
}

@article{min_2023,
  title={Large Scale Visual Food Recognition},
  author={W. Min and Z. Wang and Y. Liu and M. Luo and L. Kang and X. Wei and X. Wei and S. Jiang},
  journal={IEEE Transactions on Pattern Analysis and Machine Intelligence},
  volume={45},
  number={8},
  pages={9932--9949},
  year={2023},
  month={August},
  url={http://dx.doi.org/10.1109/TPAMI.2023.3237871}
}

@article{jia_2021,
  title     = {Scaling Up Visual and Vision-Language Representation Learning With Noisy Text Supervision},
  author    = {Jia, Chao and Yang, Yinfei and Xia, Ye and Chen, Yi-Ting and Parekh, Zarana and Pham, Hieu and Le, Quoc V. and Sung, Yun-Hsuan and Li, Zhen and Duerig, Tom},
  journal   = {Proceedings of the 38th International Conference on Machine Learning},
  volume    = {139},
  series    = {Proceedings of Machine Learning Research},
  pages     = {4904--4916},
  year      = {2021},
  month     = {July},
  note      = {{Virtual Event}},
  publisher = {PMLR},
  url       = {http://proceedings.mlr.press/v139/jia21b.html}
}

@misc{ilharco_2021,
  title        = {{OpenCLIP}},
  author       = {Ilharco, Gabriel and Wortsman, Mitchell and Wightman, Ross and Gordon, Cade and Carlini, Nicholas and Taori, Rohan and Dave, Achal and Shankar, Vaishaal and Namkoong, Hongseok and Miller, John and Hajishirzi, Hannaneh and Farhadi, Ali and Schmidt, Ludwig},
  year         = {2021},
  month        = {July},
  howpublished = {\url{https://github.com/mlfoundations/open_clip}},
  note         = {Software repository. Accessed: 2026-05-29},
  url={http://dx.doi.org/10.5281/zenodo.5143773}
}

@article{zhai_2023,
  title   = {Sigmoid Loss for Language Image Pre-Training},
  author  = {Zhai, Xiaohua and Mustafa, Basil and Kolesnikov, Alexander and Beyer, Lucas},
  journal = {Proceedings of the IEEE/CVF International Conference on Computer Vision},
  pages   = {11975--11986},
  year    = {2023},
  month   = {October},
  note    = {{Paris, France}},
  url     = {http://dx.doi.org/10.1109/ICCV51070.2023.01100}
}

@article{bossard_2014,
  title     = {{Food-101}: Mining Discriminative Components With Random Forests},
  author    = {Bossard, Lukas and Guillaumin, Matthieu and Van Gool, Luc},
  journal   = {Proceedings of the European Conference on Computer Vision},
  volume    = {8694},
  series    = {Lecture Notes in Computer Science},
  pages     = {446--461},
  year      = {2014},
  month     = {September},
  note      = {{Zurich, Switzerland}},
  publisher = {Springer},
  url       = {http://dx.doi.org/10.1007/978-3-319-10599-4_29}
}

@article{marin_2021,
  title   = {{Recipe1M+}: A Dataset for Learning Cross-Modal Embeddings for Cooking Recipes and Food Images},
  author  = {Marin, Javier and Biswas, Aritro and Ofli, Ferda and Hynes, Nicholas and Salvador, Amaia and Aytar, Yusuf and Weber, Ingmar and Torralba, Antonio},
  journal = {IEEE Transactions on Pattern Analysis and Machine Intelligence},
  volume  = {43},
  number  = {1},
  pages   = {187--203},
  year    = {2021},
  month   = {January},
  url     = {http://dx.doi.org/10.1109/TPAMI.2019.2927476}
}

@article{van_den_oord_2018,
  title   = {Representation Learning With Contrastive Predictive Coding},
  author  = {van den Oord, Aaron and Li, Yazhe and Vinyals, Oriol},
  journal = {arXiv preprint arXiv:1807.03748},
  year    = {2018},
  month   = {July},
  url     = {http://dx.doi.org/10.48550/arXiv.1807.03748}
}

@article{fang_2024,
  title   = {Data Filtering Networks},
  author  = {Fang, Alex and Jose, Albin Madappally and Jain, Amit and Schmidt, Ludwig and Toshev, Alexander and Shankar, Vaishaal},
  journal = {Proceedings of the International Conference on Learning Representations},
  year    = {2024},
  month   = {May},
  note    = {{Vienna, Austria}},
  url     = {https://openreview.net/forum?id=KAk6ngZ09F}
}

@article{chuang_2025,
  title={{Meta CLIP 2}: A Worldwide Scaling Recipe},
  author={Chuang, Yung-Sung and Li, Yang and Wang, Dong and Yeh, Ching-Feng and Lyu, Kehan and Raghavendra, Ramya and Glass, James and Huang, Lifei and Weston, Jason and Zettlemoyer, Luke and Chen, Xinlei and Liu, Zhuang and Xie, Saining and Yih, Wen-tau and Li, Shang-Wen and Xu, Hu},
  journal={arXiv preprint arXiv:2507.22062},
  year={2025},
  month={July},
  url={http://dx.doi.org/10.48550/arXiv.2507.22062}
}

@article{gao_2024,
  title   = {{CLIP-Adapter}: Better Vision-Language Models With Feature Adapters},
  author  = {Gao, Peng and Geng, Shijie and Zhang, Renrui and Ma, Teli and Fang, Rongyao and Zhang, Yongfeng and Li, Hongsheng and Qiao, Yu},
  journal = {International Journal of Computer Vision},
  volume  = {132},
  number  = {2},
  pages   = {581--595},
  year    = {2024},
  month   = {February},
  url     = {http://dx.doi.org/10.1007/s11263-023-01891-x}
}

@article{hu_2022,
  title   = {{LoRA}: Low-Rank Adaptation of Large Language Models},
  author  = {Hu, Edward J. and Shen, Yelong and Wallis, Phillip and Allen-Zhu, Zeyuan and Li, Yuanzhi and Wang, Shean and Wang, Lu and Chen, Weizhu},
  journal = {Proceedings of the International Conference on Learning Representations},
  year    = {2022},
  month   = {April},
  note    = {{Virtual Event}},
  url     = {https://openreview.net/forum?id=nZeVKeeFYf9}
}

@article{liu_2024,
  title     = {{DoRA}: Weight-Decomposed Low-Rank Adaptation},
  author    = {Liu, Shih-Yang and Wang, Chien-Yi and Yin, Hongxu and Molchanov, Pavlo and Wang, Yu-Chiang Frank and Cheng, Kwang-Ting and Chen, Min-Hung},
  journal   = {Proceedings of the 41st International Conference on Machine Learning},
  volume    = {235},
  series    = {Proceedings of Machine Learning Research},
  pages     = {32100--32121},
  year      = {2024},
  month     = {July},
  note      = {{Vienna, Austria}},
  publisher = {PMLR},
  url       = {http://proceedings.mlr.press/v235/liu24bn.html}
}

@article{yuksekgonul_2023,
  title   = {When and Why Vision-Language Models Behave Like Bags-of-Words, and What to Do About It?},
  author  = {Yuksekgonul, Mert and Bianchi, Federico and Kalluri, Pratyusha and Jurafsky, Dan and Zou, James},
  journal = {Proceedings of the International Conference on Learning Representations},
  year    = {2023},
  month   = {May},
  note    = {{Kigali, Rwanda}},
  url     = {https://openreview.net/forum?id=KRLUvxh8uaX}
}

@article{wortsman_2022,
  title   = {Robust Fine-Tuning of Zero-Shot Models},
  author  = {Wortsman, Mitchell and Ilharco, Gabriel and Kim, Jong Wook and Li, Mike and Kornblith, Simon and Roelofs, Rebecca and Lopes, Raphael Gontijo and Hajishirzi, Hannaneh and Farhadi, Ali and Namkoong, Hongseok and Schmidt, Ludwig},
  journal = {Proceedings of the IEEE/CVF Conference on Computer Vision and Pattern Recognition},
  pages   = {7959--7971},
  year    = {2022},
  month   = {June},
  note    = {{New Orleans, LA, USA}},
  url     = {http://dx.doi.org/10.1109/CVPR52688.2022.00780}
}

@misc{google_deepmind_2026,
  title        = {{Gemma 4 31B Instruct}},
  author       = {{Google DeepMind}},
  year         = {2026},
  month        = {March},
  howpublished = {\url{https://huggingface.co/google/gemma-4-31B-it}},
  note         = {Model card. Accessed: 2026-05-29}
}

@misc{radeva_2026,
  author       = {Radeva, Petia and Nagarajan, Bhalaji and Estepa, Imanol G. and Rodr{\'i}guez de Vera, Jes{\'u}s M.},
  title        = {{Dishcovery Mission II Challenge}: Where {VLM} Meets Food},
  howpublished = {\url{https://dishcoveryvlmchallenge.com/}},
  note         = {{IEEE/CVF Conference on Computer Vision and Pattern Recognition} 2026 MetaFood Workshop Challenge. Accessed: 2026-05-08},
  year         = {2026},
  month        = {February}
}

@misc{hugging_face_2026,
  title        = {{Hugging Face Hub}},
  author       = {{Hugging Face}},
  year         = {2026},
  month        = {May},
  howpublished = {\url{https://huggingface.co/docs/hub}},
  note         = {Documentation. Accessed: 2026-05-29}
}

@article{cormack_2009,
  title     = {Reciprocal Rank Fusion Outperforms {Condorcet} and Individual Rank Learning Methods},
  author    = {Cormack, Gordon V. and Clarke, Charles L. A. and Buettcher, Stefan},
  journal   = {Proceedings of the 32nd International {ACM SIGIR} Conference on Research and Development in Information Retrieval},
  pages     = {758--759},
  year      = {2009},
  month     = {July},
  note      = {{Boston, MA, USA}},
  publisher = {ACM},
  url       = {http://dx.doi.org/10.1145/1571941.1572114}
}

@article{nogueira_2019,
  title   = {Passage Re-Ranking With {BERT}},
  author  = {Nogueira, Rodrigo and Cho, Kyunghyun},
  journal = {arXiv preprint arXiv:1901.04085},
  year    = {2019},
  month   = {January},
  url     = {http://dx.doi.org/10.48550/arXiv.1901.04085}
}

@article{sun_2023,
  title   = {Is {ChatGPT} Good at Search? {Investigating} Large Language Models as Re-Ranking Agents},
  author  = {Sun, Weiwei and Yan, Lingyong and Ma, Xinyu and Wang, Shuaiqiang and Ren, Pengjie and Chen, Zhumin and Yin, Dawei and Ren, Zhaochun},
  journal = {Proceedings of the 2023 Conference on Empirical Methods in Natural Language Processing},
  pages   = {14918--14937},
  year    = {2023},
  month   = {December},
  note    = {{Singapore}},
  url     = {http://dx.doi.org/10.18653/v1/2023.emnlp-main.923}
}

\end{document}